\documentclass[runningheads]{llncs}
\usepackage[T1]{fontenc}

\usepackage{amsmath}
\usepackage{graphicx}
\graphicspath{ {./images/} }
\usepackage{xcolor}
\usepackage{caption}
\usepackage{subcaption}
\usepackage{float}
\usepackage{afterpage}

\newcommand{\etal}{\textit{et al.}}
\newcommand{\ie}{\textit{i.e. }}
\newcommand{\eg}{\textit{e.g. }}
\usepackage{amsmath}
\definecolor{cvprblue}{rgb}{0.21,0.49,0.74}
\usepackage[pagebackref,breaklinks,colorlinks,citecolor=cvprblue]{hyperref}

\usepackage[nameinlink,capitalize,noabbrev]{cleveref}
\crefname{section}{Sec.}{Secs.}
\Crefname{section}{Section}{Sections}
\Crefname{table}{Table}{Tables}
\crefname{table}{Tab.}{Tabs.}

\usepackage{caption}
\makeatletter
\renewcommand\paragraph{\@startsection{paragraph}{4}{\z@}%
                                    {3.25ex \@plus1ex \@minus.2ex}%
                                    {-1em}%
                                    {\normalfont\normalsize\bfseries}}
\makeatother

\begin{document}

\captionsetup{
  singlelinecheck=false,
  font=small,labelfont=it,belowskip=9pt,aboveskip=9pt}


\title{Multimodal Adaptive Inference for Document Image Classification with Anytime Early Exiting}
\titlerunning{Multimodal Adaptive Inference with Anytime Early Exit}
%

\author{
Omar Hamed\inst{1}, 
  Souhail Bakkali\inst{4}\orcidID{0000-0001-9383-3842} , 
    Matthew Blaschko\inst{2}\orcidID{0000-0002-2640-181X} , 
   Sien Moens\inst{2}\orcidID{0000-0002-3732-9323} , 
  Jordy Van Landeghem\inst{2,3}\orcidID{0000-0002-9838-3024} 
}

\institute{
\footnotesize Microsoft \quad
\email{omarhamed@microsoft.com}
\and
\footnotesize KU Leuven, Belgium \\
\and
\footnotesize Contract.fit, Brussels, Belgium 
\and
\footnotesize L3i - La Rochelle Université, La Rochelle, France 
}
\authorrunning{Hamed \etal}
%

%
\maketitle              

\begin{abstract}
This work addresses the need for a balanced approach between performance and efficiency in scalable production environments for visually-rich document understanding (VDU) tasks. Currently, there is a reliance on large document foundation models that offer advanced capabilities but come with a heavy computational burden. In this paper, we propose a multimodal early exit (EE) model design that incorporates various training strategies, exit layer types and placements\footnote{Code is available at \url{https://github.com/Jordy-VL/multi-modal-early-exit}}. Our goal is to achieve a Pareto-optimal balance between predictive performance and efficiency for multimodal document image classification. Through a comprehensive set of experiments, we compare our approach with traditional exit policies and showcase an improved performance-efficiency trade-off. Our multimodal EE design preserves the model's predictive capabilities, enhancing both speed and latency. This is achieved through a reduction of over 20\% in latency, while fully retaining the baseline accuracy.
This research represents the first exploration of multimodal EE design within the VDU community, highlighting as well the effectiveness of calibration in improving confidence scores for exiting at different layers. Overall, our findings contribute to practical VDU applications by enhancing both performance and efficiency.
\end{abstract}
\section{Introduction}

Thanks to the advancements in multimodal deep learning (DL) systems and hardware design, along with abundant data availability, VDU models are becoming increasingly deeper, larger, and more powerful. However, it is crucial to acknowledge that such advancements entail substantial demands on both workload and memory resources.
For production scalability, it is quintessential to investigate the performance-efficiency trade-offs for such multimodal models. Contemporary state-of-the-art (SOTA) frameworks for VDU predominantly involve multimodal models that offer advanced capabilities 
through the conventional approach of `pretrain-then-finetune' in tasks such as document image classification, named entity recognition, and document visual question answering~\cite{powalski2021going,gu2021unidoc,huang2022layoutlmv3,appalaraju2023docformerv2,wang2023docllm,blau2024gram,li2021selfdoc}. Nevertheless, substantial computational resources are necessary to fine-tune and deploy in production, a factor to consider when evaluating practical VDU applications. Improving the predictive capabilities of multimodal models requires large network parameters and computing power, albeit at the cost of efficiency.

\begin{figure}[!t]
  \centering
  \includegraphics[width=.4\linewidth]{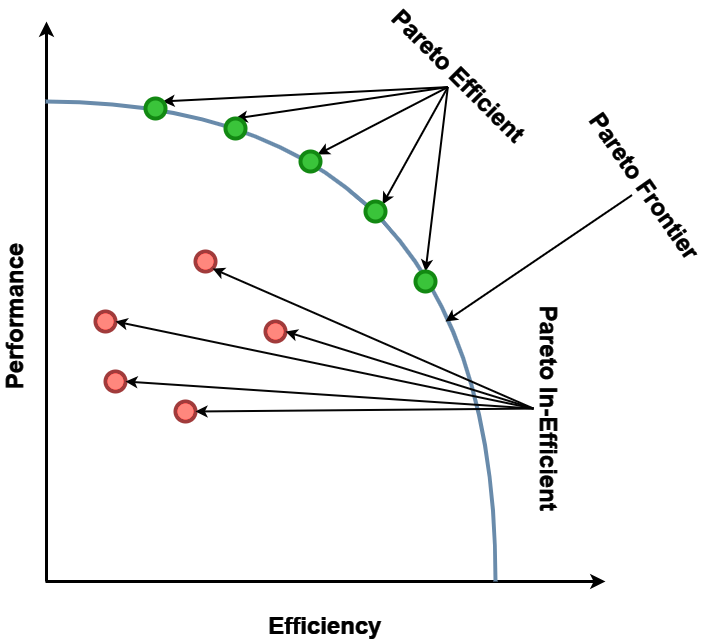}
é  \caption{Performance-efficiency trade-offs that allow comparing models for when both criteria are important, which makes any point on the Pareto frontier to be considered Pareto-efficient vs. all below the frontier.}
  \label{fig:pareto-efficiency}
\end{figure}

Conversely, to improve the efficiency of a multimodal VDU system, a compromise has to be made in its predictive capabilities. Rather than (only) chasing SOTA on task-specific performance metrics, \textbf{Pareto efficiency} \cite{liu-etal-2022-towards-efficient} ---as illustrated in \Cref{fig:pareto-efficiency}--- is proposed as an objective better aligned with VDU real-world practices, where both performance and efficiency are paramount.
Pareto efficiency allows comparing the significance of improvements on both criteria and informs threshold selection among different trade-offs and compromises that arise in scalable production settings.


When deployed in real-time industrial applications, predictive models encounter challenges in making instantaneous decisions, as evident across various applications. For instance, in the context of an information retrieval system that classifies images to facilitate efficient content indexing, the system should swiftly provide accurate classifications to support rapid and relevant information retrieval. The system must adapt to varying processing demands, enabling rapid decisions in contexts where quick identification of essential information within a document is necessary. 
Therefore, deploying image classification models on resource-constrained devices presents its own set of challenges \cite{matsubara2022split}. To the best of our knowledge, the research community has not sufficiently explored multimodal models for multimodal document image classification that can deliver high-quality results with low resource and computational requirements.

As such, in this paper, we thoroughly investigate the use of early exiting in multimodal models for document image classification, presenting a novel empirical and intuition-based thresholding scheme. We explore adaptive inference with various designs of EE strategies with a multi-exit LayoutLMv3$_{Base}$ architecture \cite{huang2022layoutlmv3}.
We assess the efficacy of our methods across a diverse set of language and vision-centric examples, by comparing the baseline with both established and proposed EE strategies. The study's experiments are aimed at delving into the intricate relationship between the complexity of documents and the feasibility of EE design to adapt inference according to instance difficulty. Based on the above, we consider the following research question: \textit{What EE architecture design would achieve SOTA Pareto-efficiency for multimodal document image classification?}

To address this question, we focus on LayoutLMv3$_{Base}$~\cite{huang2022layoutlmv3}, which has proven to be effective in various VDU tasks. During fine-tuning, we investigate different aspects of EE design such as training strategies, exit placements, and exit layer types. At inference time, we experiment with post-hoc calibration to adjust the confidence scores of the exit layers, and propose a novel heuristic EE policy for optimal thresholding to compare with both global thresholding and multi-exit thresholding. The reported results demonstrate that our multi-exit design yields substantial improvements in terms of speed and latency, while maintaining the majority of the model's predictive capabilities. This underscores the practicality and effectiveness of our proposed methods in achieving notable efficiency gains without compromising predictive accuracy.

\section{Related work}
\paragraph{Dynamic Neural Networks with Adaptive Early-Exit Inference.} Dynamic Neural Networks \cite{xu2022survey,xing2020early,zhou2020bert} are an enhanced type of neural networks that can dynamically adjust their computational path based on the input, resulting in significant improvements in inference time. Multiple research efforts investigate how to effectively design early-exit networks. DeeBERT \cite{tishby2015deep} updates BERT architecture by inserting intermediate classifiers between the encoder layers, then jointly fine-tunes them with the original classification layer while freezing the model backbone. This method achieves up to 40\% inference time savings with minimal loss of accuracy on GLUE tasks. In \cite{geng2021romebert}, RomeBERT improved on DeeBERT by introducing gradient regularization and self distillation. Gradient regularization allows the training of the model backbone whereas self-distillation allows the use of the predictions of the final layer as soft targets for the off-ramps, and minimizes the KL-divergence between them to fix the significant drop in intermediate-classifiers performance. BERxiT \cite{xin2021berxit}, improves on the existing EE architectures and extends the early exiting design for regression tasks by introducing a learning-to-exit (LTE) module. The LTE module and the intermediate classifiers are set up simultaneously and shared by all layers. The whole model, along with the LTE module, is then trained end-to-end.

\paragraph{Training Strategies for Dynamic Inference.} Different training strategies have shown their effectiveness in training early-exit networks. In \cite{laskaridis2021adaptive}, the authors discuss two strategies. The first strategy is end-to-end training, which minimizes a single loss function that combines the losses of the intermediate classifiers and the final classifier. The second strategy is training of intermediate classifiers, which consists of two stages: in the first stage, the backbone model is trained without any extra classifiers; in the second stage, the backbone weights are frozen, and the intermediate classifiers are added and trained separately. Additionally, \cite{xin2021berxit} presents a novel fine-tuning strategy named Alternating. It alternates between two objectives for each iteration, based on whether the iteration is odd or even. For odd iterations, it only updates the weights of the backbone model, while during even iterations, it updates only the intermediate classifiers.

 \paragraph{Inference Strategies.} There are different inference strategies that determine the model's exiting behavior. Inference strategies are composed of two main points: exit criterion and exit policy. The exit criterion includes confidence scoring functions (CSFs) that allow the model to exit early based on the input difficulty and the quality of the predictions. Meanwhile, the exit policy is associated with thresholding the early exiting branches.
One of the common CSFs is maximum confidence \cite{laskaridis2021adaptive,huang2017multi}, where the model exits at the first branch that satisfies a pre-defined confidence threshold. Another approach is the entropy-based criterion \cite{teerapittayanon2016branchynet}, where the model exits at the branch with the least entropy value. DynamicDet \cite{tang2023you} proposes an adaptive router for image classification tasks, using an exit criterion based on the detection losses of each exit classifier, which are employed to compute a reward signal for the adaptive router. DVT \cite{wang2021not} presents an architecture with a series of Transformers with varying token numbers. These are activated one by one at inference time, depending on the confidence score of each classifier in the series. Additionally, Zhou \etal \cite{zhou2020bert} propose EE when multiple consecutive classifiers agree on the same prediction.
Regarding the exit policy, the majority of research adopts a greedy approach, utilizing a single threshold for all exit classifiers \cite{baccarelli2020optimized,berestizshevsky2019dynamically,li2019improved}. In contrast, alternative methods aim to determine thresholds more effectively. For example, EENET \cite{ilhan2024adaptive} employs a scheduling architecture to learn an optimal early-exit policy. Similarly, \cite{zhang2019scan} utilizes genetic algorithms to learn optimal thresholds, while \cite{jayakodi2018trading} applies Bayesian optimization for the same purpose.

While prior research has focused on unimodal architectures, this paper breaks new ground by introducing EE network design for multimodal DU tasks. The most comparable work is DynMM \cite{xue2023dynamic}; however, it employs a different approach, incorporating data fusion from diverse modalities.
Our work analyzes how different exit classifiers (such as gates and ramps), diverse training strategies, various exit placements, and different exit policies impact the multimodal EE architecture. The study also explores the substantial performance enhancements achieved through the calibration of exit classifiers. Moreover, we present a novel and empirical exit policy heuristic that surpasses traditional methods without adding any extra computational burden for finding optimal thresholds.

\section{Methodology}

This work focuses on optimizing the use of LayoutLMv3$_{Base}$ across modalities through an EE network design. This design allows premature termination from the text-modality layer, visual modality layer, or their combined output, after modality fusion or different encodings. In multimodal architectures, dedicated layers for each modality facilitate tailored EEs, optimizing inference based on efficiency. In LayoutLMv3$_{Base}$, exiting from the vision modality is particularly efficient, avoiding the costly OCR step for text. Drawing on the information bottleneck theory \cite{tishby2015deep}, deeper layers capture more information, enabling the model to learn intricate patterns for accurate predictions. Firstly, we investigate various exit placements by adding intermediate classifiers at different parts of the network, and exploring both gates \cite{xin2021berxit} and ramps \cite{geng2021romebert} as exit layer types.
Secondly, to optimize a multi-exit network, we explore various fine-tuning strategies such as \textit{subgraph weighting},\textit{exit entropy regularization}. Finally, we compare different exiting and inference policies, such as \textit{global thresholding and guided random multi-exit threshold}, and propose a new heuristic exit policy.

\subsection{EE Design}
Given a multimodal input-output pair $(x, y)$ and a pre-trained multimodal architecture with $L$ layers, organized according to a computational graph $G$ (allowing for potential parallel computation), the goal is to obtain a parameterized mapping $f(\cdot; \theta): \mathcal{X} \to \mathcal{Y}$, where $\mathcal{Y} = [K]$ for $K$ mutually exclusive classes. This is typically obtained by minimizing a loss function (\eg, cross-entropy (CE)) over the last layer output (logits), $\mathcal{L}_{\mathrm{CE}}(Y, f(X;\theta_L)$, where $\theta_L$ holds all (trainable) parameters until the $L$-th layer. An EE model extends $f$ by introducing $B$ additional intermediate classifiers, which (for layer $l \in [L]$) can be either a \textit{gate} $g: f(X;\theta_{l-1} \to {0,1})$ passing on the intermediate representation to the final classifier, or a \textit{ramp} $r(\cdot; \theta): f(X;\theta_{l-1}) \to [K]$ directly making a final prediction.

\subsection{Fine-tuning Strategies} 
\subsubsection{Training Configuration.}
Exploring the trade-offs in EE design, we implement three distinct training strategies that offer different approaches for updating the weights of $B$ subgraphs in a multi-exit network, with each \textit{subgraph} $G_b \subset G$ incorporating all parameters ($\theta_{G_b}$) used up to the corresponding exit branch. The most straightforward strategy, known as multi-exit loss, involves combining the losses for each subgraph with the classifier loss:
\vspace{-\parskip} 
\begin{align}
    \mathcal{L}_{\mathrm{CE}}(Y, f(X;\theta)) + \sum_{b=1}^{B} \mathcal{L}_{\mathrm{CE}}(Y, f(X;\theta_{G_b})) 
    \label{eq:equation_1}
\end{align}
This objective assumes uniform weights for all $B$ loss terms and makes no assumptions about the importance of exits compared to the final classifier. Other training strategies enhance \Cref{eq:equation_1} by i) restricting loss updates to parameters within the scope of the exit branch, ii) incorporating a hyperparameter ($\gamma \in [0,1]$) to regulate the influence of all exits versus the classifier \cite{geng2021romebert}, and iii) introducing an entropy regularization term over all exit losses to assign weights based on normalized uncertainty relative to other exits.


\begin{itemize}
    \item \textbf{Subgraph Weighting}: This training strategy implements $\gamma$ to tradeoff between exit classifiers and the final classifier. Additionally, each exit weight is based on the relative number of parameters used, with the inverse weighting reinforcing the importance of earlier exits.
    \vspace{-\parskip} 
    \begin{align}
    (1-\gamma) \underbrace{\color{black}\mathcal{L}_{\mathrm{CE}}(Y, f(X;\theta))}_{\text{Final classifier loss}} + \gamma \underbrace{\sum_{b=1}^B w_{b} \mathcal{L}_{\mathrm{CE}}(Y, f(X;\theta_{G_b}))}_{\text{Exit classifiers loss}} 
        \label{eq:equation_4}
    \end{align}


where $w_{b}$ is the weight assigned to the b-th exit (\eg using 5\% of parameters would give a weight of 20), and $\theta_{G_b}$ is the set of parameters up to the b-th layer.

    \item \textbf{Exit Entropy Regularization}: This strategy relies on an entropy regularization term \cite{araujo2022entropy} to discourage branches with high uncertainty. 
    Intuitively, exits with high entropy respective to all other exits will receive less importance than exits with low entropy, which exhibit high certainty in their prediction. 
    The entropy regularization term measures the mutual information between each branch and its target output, and imposes a penalty on branches with lower mutual information. 
    
    \vspace{-\parskip} 
    \begin{align}
        (1-\gamma) \underbrace{\color{black}\mathcal{L}_{\mathrm{CE}}(Y, f(X;\theta))}_{\text{Final classifier loss}} + \gamma \underbrace{\sum_{b=1}^B \lambda_b \mathcal{L}_{\mathrm{CE}}(Y, f(X;\theta_{G_b}))}_{\text{Exit classifiers loss}} 
        \label{eq:equation_4}
    \end{align}

    where $\lambda_{b}$ is the regularization coefficient for the b-th exit, which arises from the calculation of the exit layer entropy $H(x,\cdot)$, normalized over all other exits, and inversed, $\mathbf{\lambda} = 1-\mathrm{softmax}(H(Y;f(X;\theta_{G})))$.

    \item \textbf{Weighted - Entropy Regularization}: This strategy integrates both subgraph weighting and entropy regularization strategies, by using a weighted loss function and an entropy regularization term for each exit. When applying this joint strategy, we set $\gamma=0.5$ to ensure weighting stability.

    
    

\end{itemize}

\subsubsection{Exits Placement Configuration.} We conduct experiments with diverse exit branch configurations throughout the network to assess their impact on model performance. Given the exponential increase in possible combinations with network depth, we choose six exit placement configurations capable of capturing distinct model aspects. Subsequently, each of these placement configurations undergoes training using the aforementioned strategies. The exit configurations are as follows:

\begin{itemize} 
    \item \textbf{Independent-All}: Each modality (vision and language) has its own exit branch, along with an exit branch at every encoder layer. This results in 14 exits in total.
    
    \item \textbf{Concat-All}: A single exit branch is positioned after the concatenation layer, along with an exit branch at every encoder layer. This results in 13 exits in total.
    
    \item \textbf{Concat-Single}: A single exit branch is positioned after the concatenation layer, and another exit branch is positioned after the 6th encoder layer. This results in 2 exits in total.
    
    \item \textbf{Concat-Quarter}: Four exit branches are placed at the 1st, 4th, 8th, and 12th encoder layers, in addition to the exit branch after the concatenation layer. This results in 5 exits in total.
    
    \item \textbf{Concat-Alternate}: Four exit branches are located at the 2nd, 5th, 9th, and 11th encoder layers, in addition to the exit branch after the concatenation layer. This also results in 5 exits in total.
\end{itemize}

\subsubsection{Exit Classifiers Type.} Moreover, we investigate the impact of various exit classifiers on model performance. For each combination of a training strategy and an exit placement configuration, we employ either ramps (\ie softmax activation) or gates (\ie sigmoid activation) as the exit classifier, resulting in a base total of 32 experiments (without any hyperparameter tuning). A gate takes the intermediate representation $f(X;\theta_{b-1})$ as input and produces after a sigmoid activation a probability $p_{b,i}$ for exiting. The gate is trained by minimizing a binary cross-entropy loss function, expressed as follows:

\vspace{-\baselineskip} 
\begin{align}
    \mathcal{L}_{\mathrm{gate}}(Y, f(X;\theta_{b-1})) = -\frac{1}{N}\sum_{i=1}^N y_i \log p_{b,i} + (1-y_i) \log (1-p_{b,i})
    \label{eq:equation_2}
\end{align}
Where N represents the number of samples, $y_{i}$ denotes the ground truth label (0 or 1) for the i-th sample, and $p_{b,i}$ represents the output probability of the gate for the i-th sample at exit layer $b$. 

Conversely, a ramp serves as a multi-class classifier producing a probability distribution across $K$ classes to indicate the predicted class label for the input. It involves a softmax function, which takes the intermediate representation $f(X;\theta_{b-1})$ as input and generates a vector $p_{b,i}$ containing probabilities for each class. The training of a ramp is accomplished by minimizing a cross-entropy loss function, as follows:
\vspace{-\baselineskip} 
\begin{align}
    \mathcal{L}_{\mathrm{ramp}}(Y, f(X;\theta_{b-1})) = -\frac{1}{N}\sum_{i=1}^N \sum_{k=1}^K y_{i,k} \log p_{b,i,k}
    \label{eq:equation_3}
\end{align}
Where N represents the number of samples, $y_{i,k}$ denotes the ground truth label (0 or 1) for the i-th sample and the k-th class, and $p_{b,i,k}$ represents the output probability of the ramp for the i-th sample and the k-th class at exit layer $b$.

\subsection{Inference Strategies}

\subsubsection{Exit criterion.} In this work, we opt for the Maximum Softmax Probability (MSP), shown in \Cref{eq:equation_13} as the exit criterion. We were motivated by the minimal difference observed compared to using alternative strategies such as entropy. Additionally, MSP, being an L2 metric, produces values within the range of 0 to 1, simplifying the determination of exit thresholds.
\vspace{-\parskip}
\begin{align}\displaystyle
     p_{b} \approx \mathrm{MSP}(x;\theta_{G_b}) = \max_{k \in K} 
     \frac{\exp(z_{b,k})}{\sum_{c=1}^{K}{\exp{(z_{b,c})}}}, \text{  } b \in [1, B-1],
     \label{eq:equation_13}
\end{align} where $z_{b}$ denotes the raw logits at layer $b$. The exit criterion should indicate the 'probability' of correctly exiting, as the MSP should be an approximation of.

\subsubsection{Calibration.} Initiating our evaluation, we calibrate the exit logits of each layer using a validation set. Employing temperature scaling, we aim to determine the optimal temperature $T$ that minimizes cross-entropy. Subsequently, this optimal temperature value is applied to smoothen the logits $z$ at each exit classifier, as outlined in \Cref{eq:equation_7}:
\vspace{-\parskip}
\begin{align}
    p_{b,k}  = \frac{\exp({\frac{z_{b,k}}{T}})}{\sum_{c=1}^{K}{\exp{(\frac{z_{b,c}}{T})}}}, \text{  } b \in [1, B-1],
    \label{eq:equation_7}
\end{align}
where $T$ represents the temperature scaling factor, and $p_{b,k}$ represents the calibrated probabilities at exit layer $b$ for class $k$.

\subsubsection{Global Thresholding.} Exits are activated when their confidence value exceeds a certain threshold. However, the challenge comes in choosing this threshold for each exit. A common approach is to assign a single global threshold for all exits as shown in \Cref{eq:equation_8}. Despite that the global threshold approach could be a simple-to-setup solution that can achieve some computational reduction gains, it is still suboptimal.
\vspace{-\parskip}
\begin{align}
    \hat{y} = \begin{cases}
    f_{b}(x;\theta) & \text{if } p_b \geq \tau, b \in [1, B-1] \land \tau \in ]0, 1[ \\
    f_L(x;\theta) & \text{otherwise}
    \end{cases}
    \label{eq:equation_8}
\end{align}
where $x$ is the input, $\hat{y}$ is the output prediction, $f_{b}$ and $p_{b}$ are the output prediction and output probability of the b-th intermediate classifier respectively, $\tau$ is the global threshold, and $B$ is the number of exit classifiers.

\subsubsection{Multi-exit Thresholding.} exploits more degrees of freedom where each layer has its own threshold as shown in \Cref{eq:equation_9}. The best thresholds would be found by a brute force method, but this is impractical because the number of possible thresholds grows exponentially with the number of exits. 
We use a heuristic random subsampling method and limit our experiments to $10e+6$ random combinations of thresholds for each exit. To find the potential thresholds for each exit, this method applies a simple heuristic to obtain data-driven thresholds: it computes the MSPs for each exit and then selects the $i^{th}$ percentile of the MSP vector as a potential threshold. Different 10th percentiles are randomly selected per exit and combined with those of other exits to create diverse multi-threshold sets.

\vspace{-\parskip}
\begin{align}
    \hat{y} = \begin{cases}
    f_b(x;\theta) & \text{if } p_b \geq \tau_{b} \text{ where } b \in [1, B-1] \land \tau_{b} \in ]0, 1[ \\
    f_L(x;\theta) & \text{otherwise}
    \end{cases}
    \label{eq:equation_9}
\end{align}
where $\tau_{b}$ is the selected threshold for a layer $b$.

\subsubsection{Proposed Thresholding Heuristic.} Given the shortcomings of both the global and randomized multi-exit thresholding, we propose a dynamic thresholding heuristic to serve as a well-rounded approximation. The proposed formula sets a different threshold for each layer, depending on the tradeoff between accuracy and calibration error at that layer. A lower threshold value indicates a better balance between accuracy and calibration, while a higher value indicates a worse balance. A threshold value of 0 means that the accuracy and calibration error are equal, implying perfect calibration. A threshold value of 1 means that the accuracy is zero, implying complete inaccuracy. The proposed heuristic can be expressed as follows:

\vspace{-\baselineskip}
\begin{align}
    T_{b} = 1 - \frac{\mathrm{ACC}_{b}}{\mathrm{ECE}_{b}} \text{ where } b \in [1, B-1]
    \label{eq:equation_10}
\end{align}
where $\mathrm{ACC}_{b}$ and $ECE_{b}$ are the post-calibration accuracy and expected calibration error at exit layer $b$ respectively.

To measure the discrepancy between the average confidence and the average accuracy of the predictions, we use the expected calibration error (ECE), which is computed over $J$ bins as follows:
\begin{align}
    \text{ECE} = \sum_{j}^J \frac{\mathcal{B}_{j}}{N} \vert{\mathrm{ACC}(\mathcal{B}_{j}) - \mathrm{MSP}(\mathcal{B}_{j})\vert}
    \label{eq:equation_11}
\end{align}
where $\mathrm{ACC}(\mathcal{B}_{j})$ and $\mathrm{MSP}(\mathcal{B}_{j})$ represent the average probability and the expected confidence for bin $j$, respectively. 

A min-max normalization is applied to the potential thresholds for each layer. A tuneable parameter $\varepsilon$ is introduced to the min-max normalization, which avoids the thresholds from being exactly 0 or 1 and controls the gap between them. The normalization step is as in \Cref{eq:equation_12}:
\begin{align}
    \overrightarrow{\tau}_{normalized} = \frac{{(\overrightarrow{\tau})} - (\min(\overrightarrow{\tau}) - \varepsilon)}{(\max(\overrightarrow{\tau}) + \varepsilon) - (\min(\overrightarrow{\tau}) - \varepsilon)}
    \label{eq:equation_12}
\end{align}
where $\overrightarrow{\tau}$ is a vector for all layers thresholds [$\tau_{1}$, $\tau_{2}$, ..., $\tau_{B-1}$] and $\varepsilon$ is a tunable parameter to control the absolute min and max values.

\section{Experiments and Results}
\paragraph{Experimental Setup.} To evaluate the effectiveness of our proposed EE design, we use a subset\footnote{\url{https://huggingface.co/datasets/jordyvl/rvl_cdip_100_examples_per_class}} of the Ryerson Vision Lab Complex Document Information Processing (RVL-CDIP) benchmark dataset~\cite{harley2015evaluation}. It consists of 800 samples for training, 400 samples for validation, and 400 samples for test sets. The samples are balanced over the 16 classes : \textit{advertisement, budget, email, file folder, form, handwritten, invoice, letter, memo, news article, presentation, questionnaire, resume, scientific publication, scientific report, specification}. The motivation for limiting experiments on a subset of data comes on the one hand, from the need to explore the exponential size EE design space, while on the other hand, real-word settings typically do not provide 25K samples per document type, thus imposing real-world constraints. 

We apply easyOCR to extract the text within document images along with the 2D bounding box coordinates. We use LayoutLMv3$_{Base}$ as the baseline model and evaluate its performance based on both accuracy and efficiency, with latency as a measure. Latency is then calculated as a function of the exit location relative to the final classifier, assuming identical parameters and complexity for all exit classifiers. We start our experiments by re-implementing LayoutLMv3$_{Base}$ on the RVL-CDIP subset., and denote the result as LayoutLMv3$_{Base}^\dagger$. 
LayoutLMv3$_{Base}^\dagger$ yields and accuracy of 80.75\% on document image classification (\ie with no exits), which is used as a benchmark for the conducted experiments. For fine-tuning, AdamW is used as an optimizer with a batch size of 2 and a learning rate of 1e-4 for 20 epochs.
\begin{figure}[!t]
  \centering
  \includegraphics[width=\linewidth]{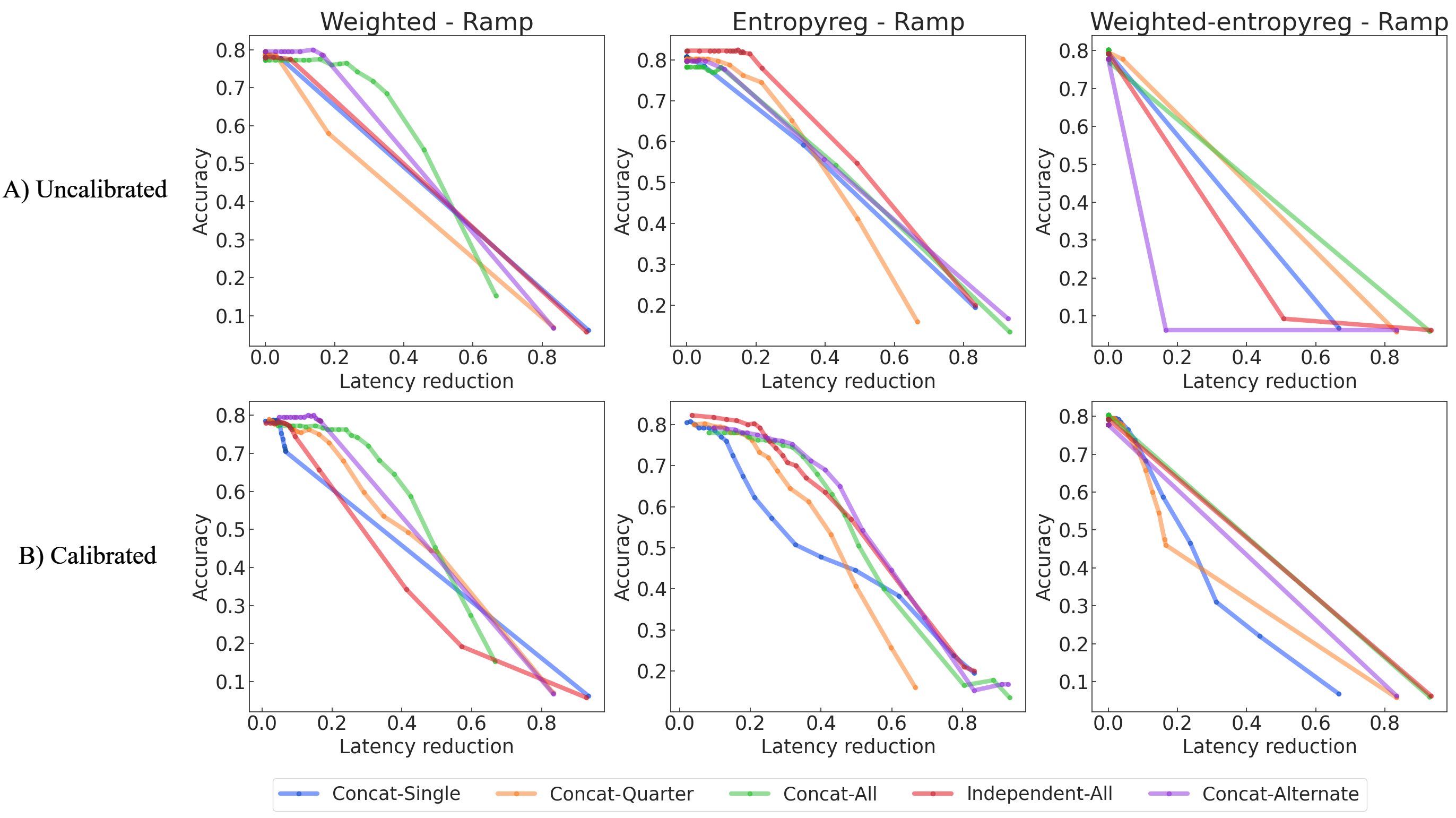}
  \caption{Effect of calibration for Ramp exits and training strategy.}
  \label{fig:calibration_ramps}
\end{figure}

\paragraph{Effect of Calibration for Ramp Exits.} In \Cref{fig:calibration_ramps}, the comparison between calibrated and uncalibrated exits for ramps is depicted, focusing on the tradeoffs between performance and efficiency. The x-axis illustrates latency reduction, while the y-axis represents accuracy. The graphs span thresholds from 0 to 1 with a step of 0.05, revealing the impact of calibration on the tradeoff between performance and efficiency across various training methods and the ramp exit classifier in global thresholding. The results show that calibration facilitates more adaptable thresholding in most scenarios, whereas the uncalibrated version exhibits constraints, with latency improvements leveling off at relatively low thresholds compared to the calibrated version. For example, we observe that the calibrated versions of \textit{Weighted - Ramp} and \textit{Entropyreg - Ramp} have better performance-efficiency tradeoffs than the uncalibrated versions for all models. The \textit{Weighted-entropyreg - Ramp} performs worse than other training strategies, requiring better tuning of the multiple weightings.



\begin{figure}[th]
  \centering
  \includegraphics[width=\linewidth]{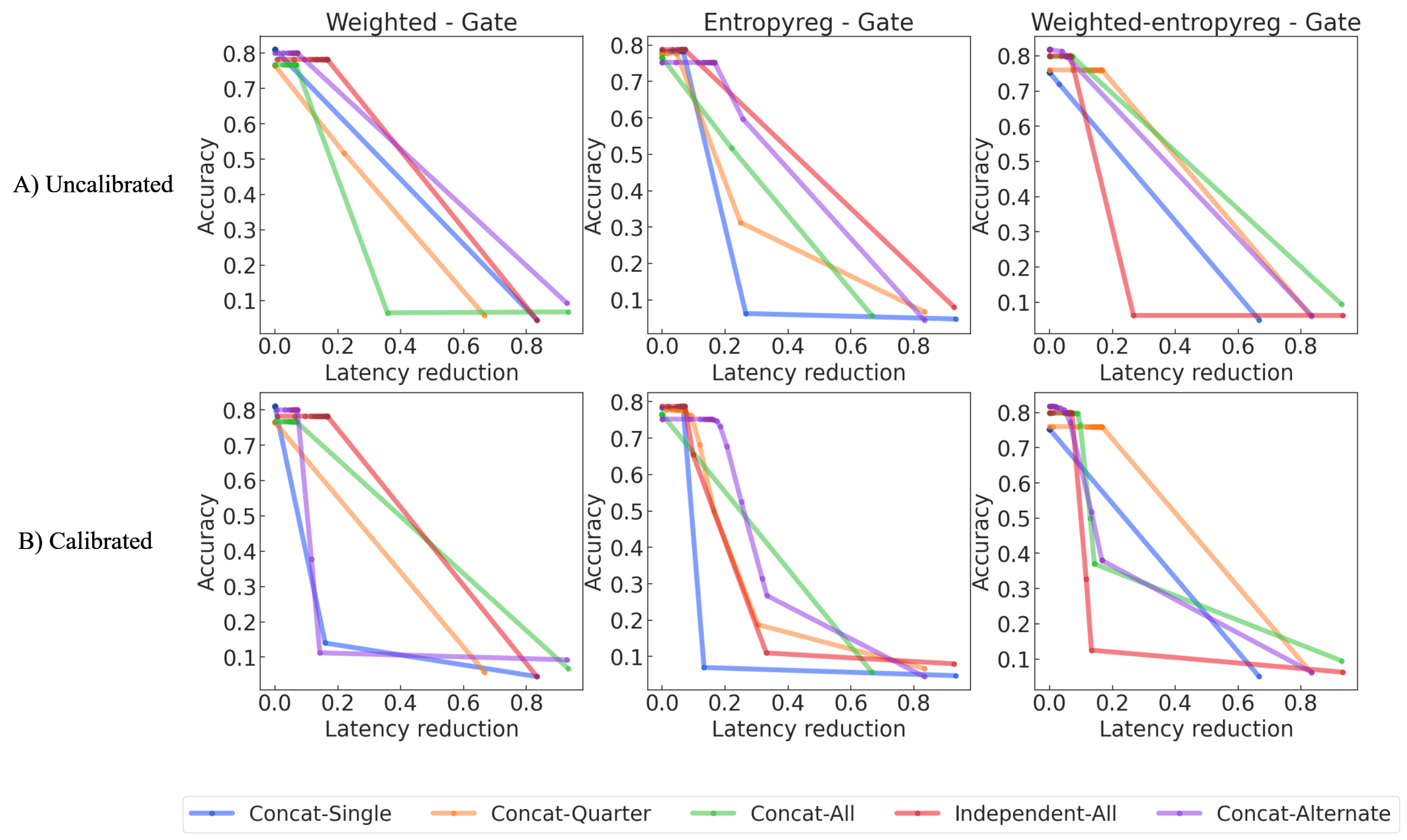}
  \caption{Effect of calibration for Gate exits and training strategy}
  \label{fig:calibration_gates}
\end{figure}
\paragraph{Effect of Calibration for Gate Exits.} In contrast to the ramp exit type, the gate exit type exhibits a distinct Pareto-efficiency trend, as depicted in \Cref{fig:calibration_gates}. Relative to ramps, gates demonstrate less pronounced latency reduction at lower thresholds. 
Despite this, calibration consistently enhances trade-offs in most scenarios compared to their uncalibrated counterparts. Interestingly, specific configurations perform better with ramps than with gates, as seen in the case of the \textit{Concat-single} configuration, which exhibits limited latency reduction with gates but higher efficiency with ramps. In addition, different training methods impact how well a specific model configuration performs. The \textit{weighted} training method seems to achieve a higher Pareto-efficiency tradeoff, especially for the calibrated version, than the other training methods. For instance, the \textit{Concat-All} variant can balance performance and efficiency better with the \textit{weighted} method than the \textit{weighted-entropyreg} method.





\begin{figure}[!t]
  \centering
    \makebox[1\textwidth][c]{
  \includegraphics[width=1.1\linewidth]{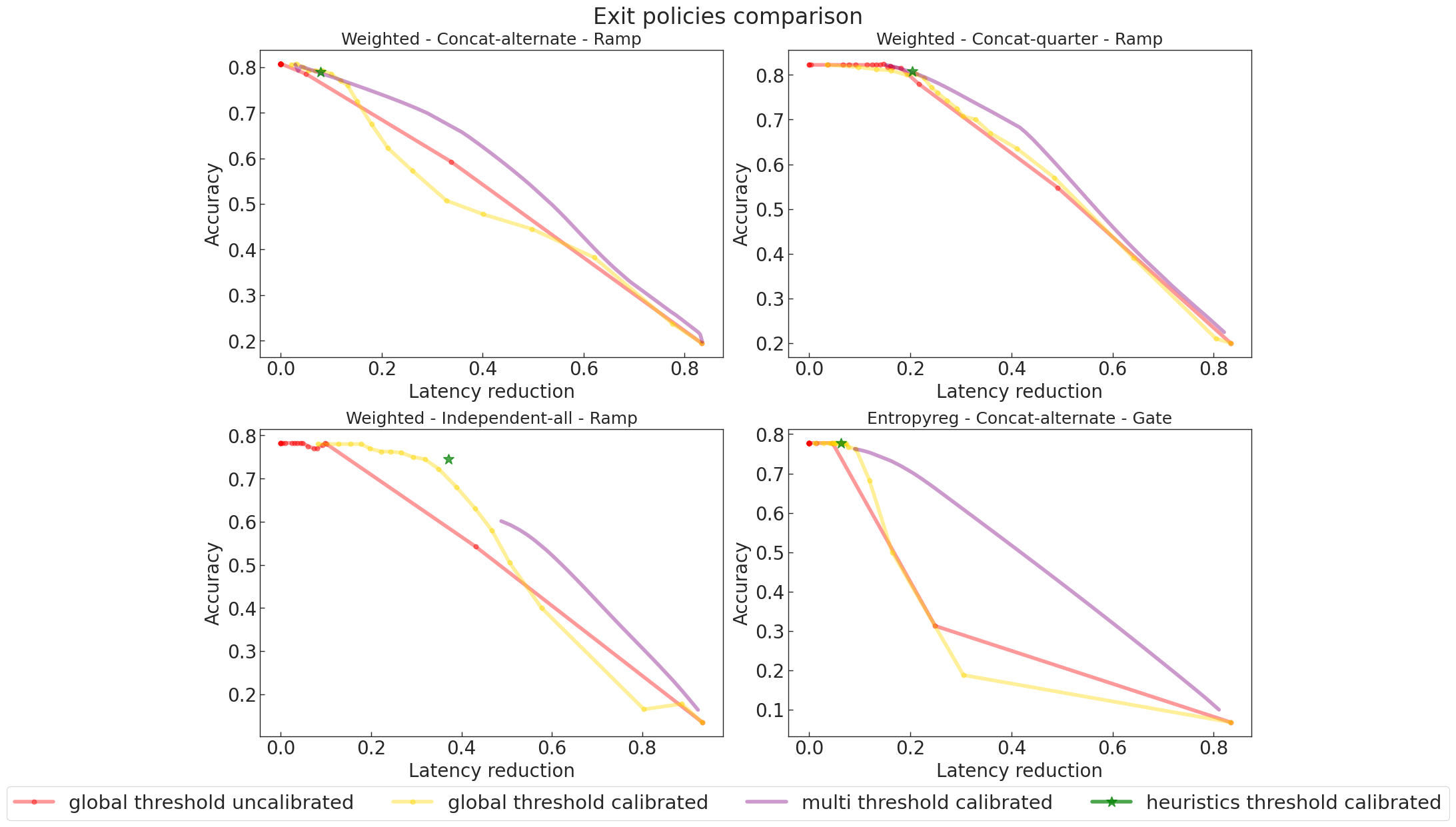}}
  \caption{A comparison between the different exit policies, varying calibration.}
  \label{fig:full_comparison}
\end{figure}

\paragraph{Exit Policies Comparison.} \Cref{fig:full_comparison} compares the performance of different exit policies (\textit{global threshold uncalibrated}, \textit{global threshold calibrated},  \textit{multi-exit thresholding}, and \textit{proposed thresholding heuristic}) for a subset of model configurations. These configurations are denoted as \textit{Weighted---Concat-alternate---Ramp}, \textit{Weighted---Concat-quarter---Ramp}, \textit{Weighted---Independent-all---Ramp}, and \\ 
\textit{Entropyreg--- Concat-alternate ---Gate}.  
The proposed heuristic can achieve a similar or better tradeoff than global or multi thresholding on calibrated exits without requiring extra tuning. For example, the heuristic method has higher accuracy and comparable latency reduction to global thresholding in the \textit{weighted --- Independent-all --- Ramp} configuration. However, the heuristic method is less effective than global thresholding in the \textit{Entropyreg --- Concat --- alternate --- Gate} configuration in terms of latency improvement, yet it yields better accuracy. Nevertheless, it is evident that the heuristic method can offer a competitive tradeoff without requiring tuning or searching in the multi-threshold space. The optimal inference strategy depends on the use case and the preference for computation efficiency or accuracy. Some may take advantage of the many options of multi-thresholding, while others may simply resort to the heuristic method. \\
Taking the \textit{weighted --- Concat-quarter --- Ramp} as an example, it demonstrates better performance than the baseline model. In the heuristic approach, with an accuracy of 80.75\%, the associated latency reduction is 20\%. Comparatively, employing global thresholding showcases varying latencies for different accuracy levels: at 81\%, the latency is 16.16\%; at 81.25\%, it drops to 13.29\%, and further decreases to 9\% at an accuracy of 81.75\%. This analysis underscores the nuanced trade-offs between accuracy and latency under different thresholding strategies.

\begin{figure}[!t]
  \centering
  \includegraphics[width=\linewidth]{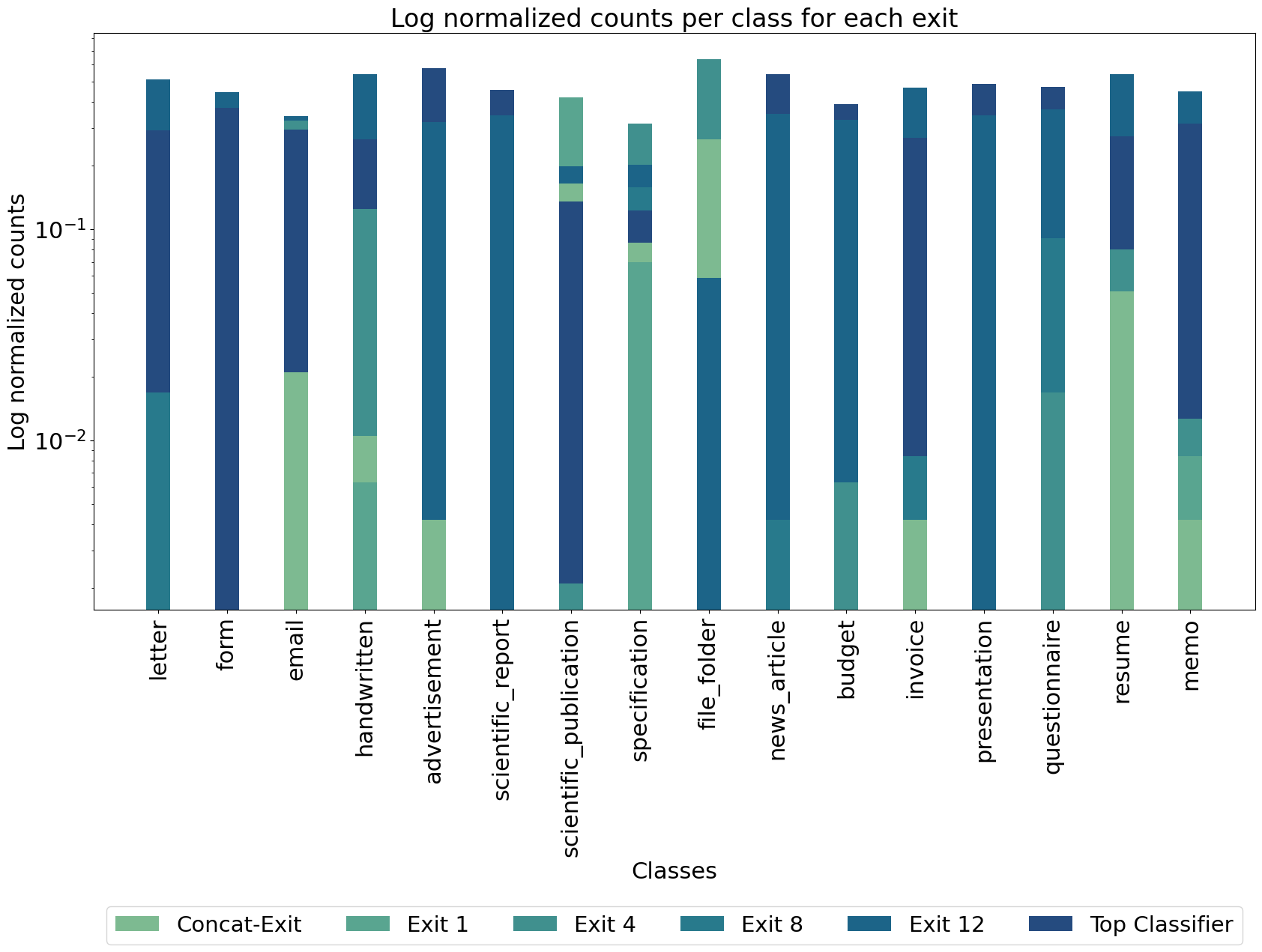}
  \caption{Exit patterns for \textit{Weighted — Concat-quarter — Ramp} configuration on RVL-CDIP dataset classes, showing the exit behaviour for different intermediate classifiers per class}
  \label{fig:exit_patterns}
\end{figure}

\begin{figure}[!t]
    \centering
    \includegraphics[width=.8\textwidth]{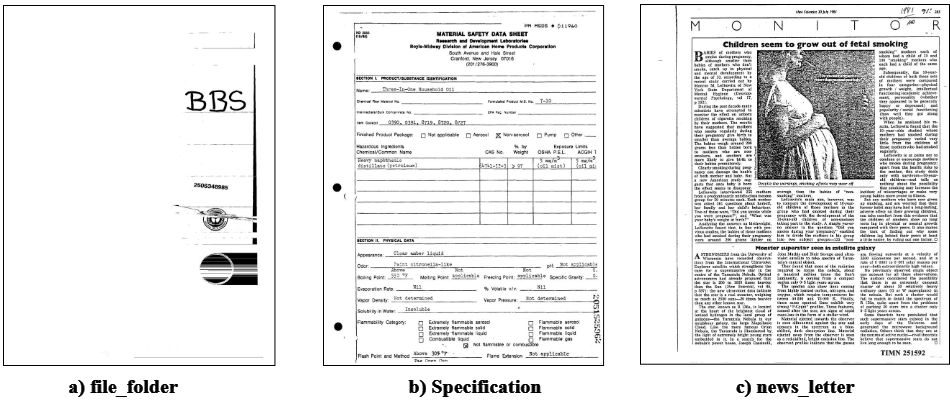}
    \caption{Samples for different classes with different difficulty and exited at different places in the network}
    \label{fig:qualitative_analysis}
\end{figure}

\paragraph{Exit Patterns.} Furthermore, we carried out a qualitative analysis to identify patterns about which classes tend to be predicted correctly and allow the model to exit at a relatively early stage compared to the final top classifier. 
We used the \textit{Weighted --- Concat-quarter --- Ramp} model as an example, as shown in \Cref{fig:exit_patterns}. The figure shows log-scaled normalized counts of how often each class is exited at a given intermediate classifier. It is clear from \cref{fig:exit_patterns} that some classes, such as \textit{file\_folder}, \textit{specification}, and \textit{scientific publication}, are more likely to be exited correctly at relatively earlier stages, indicating that these classes are somewhat easier to classify than others. On the other hand, classes such as \textit{news\_article} and \textit{scientific\_report} are more likely to be exited at relatively later stages or even at the top classifier, indicating that they require more complex and rich features for the model to learn to be accurately predicted. We note that the exact exit classifier may differ depending on the model configuration, due to changes in exit placements and the number of classifiers. However, interestingly, the exit patterns for classes are remarkably consistent across different models. \cref{fig:qualitative_analysis} illustrates different samples that are correctly classified at different stages of the network. As for the \textit{file\_folder} category, it is the easiest to identify among the other classes because it primarily comprises a white page with few texts located at the bottom right corner of the document image. This type of content does not require processing through all encoders to be accurately predicted. In contrast, the \textit{specification} category poses a slightly greater challenge due to its larger amount of text and intricate layout. This includes the presence of substantial structured tables, a characteristic shared with other classes such as \textit{form}, \textit{questionnaire}, and \textit{scientific\_publication}. These variations contribute to the complexity of inter-class variability, which presents a notable challenge in multimodal document image classification. Moreover, the \textit{news\_article} category poses the greatest challenge in classification, as it is frequently mistaken for the \textit{advertisement} category. This confusion arises because both categories often feature similar visual elements, including headlines, images, and text blocks. Consequently, classification models struggle to distinguish between them based solely on visual or text features. As a result, document images typically need to be processed through more layers of the network to achieve accurate predictions.

\section{Conclusion}

This research represents the first exploration of multimodal early exiting in the document understanding community. The research studies how to achieve a controllable balance between efficiency and performance in VDU tasks. The EE design incorporates various training strategies, exit layer types, and exit placements to achieve Pareto-optimality. Our experiments demonstrate the importance of calibrating the exit branches to significantly improve the possible tradeoffs for flexible tuning based on the use case. Additionally, different exit thresholding policies were compared, in which we observed improvements in performance-efficiency trade-offs with the proposed thresholding approaches, with a reduction of over 20\% in latency while maintaining a minimal loss in accuracy when weighted subgraphs and ramps are used as training strategy and exit type, respectively. 
Overall, the proposed methodology opens avenues for further research in adaptive inference for multimodal tasks, offering a promising direction for scalable VDU applications.

\paragraph{Limitations.} 
While our research has yielded promising results, it is essential to acknowledge its limitations. 
Our experiments were focused on a classification task that is discriminatively modeled, \ie a limited output space respective to the number of classes, whereas EE design for generative models remains to be explored, \eg large language models with outputs requiring autoregressive decoding and early exiting at different timesteps. 
Additionally, we did not study the effect of combining multiple inference optimization approaches such as distillation, quantization, and pruning with the current EE design. 
As the experiments were conducted on a smaller subset, it would be interesting to see if the results scale with larger training data or with larger architectures.
These unexplored aspects present opportunities for future research to enhance the comprehensiveness of our findings and provide a more nuanced understanding of the proposed approach.

\clearpage

\newpage
\bibliographystyle{splncs04}
\bibliography{main}

\end{document}